\documentclass[12pt]{article}
\begin{document}

\title{On the complexity of learning a language: \\An improvement of Block's algorithm}

\author{Eric Werner 
\thanks{Balliol Graduate Centre, Oxford Advanced Research Foundation (http://oarf.org).
\copyright Eric Werner 2012.  All rights reserved. }\\ \\
University of Oxford\\
Department of Physiology, Anatomy and Genetics, \\
and Department of Computer Science, \\
Le Gros Clark Building, 
South Parks Road, 
Oxford OX1 3QX  \\
email:  eric.werner@dpag.ox.ac.uk\\
Website: http://ericwerner.com
}

\date{ } 

\maketitle

\begin{abstract}  

Language learning is thought to be a highly complex process. One of the
hurdles in learning a language is to learn the rules of syntax of the
language. Rules of syntax are often ordered in that before one rule can applied
one must apply another. It has been thought that to learn the
order of $n$ rules one must go through all $n!$ permutations.  Thus to
learn the order of $27$ rules would require $27!$ steps or
$1.08889x10^{28}$ steps.  This number is much greater than the number of seconds
since the beginning of the universe!   In an insightful analysis the
linguist Block ([Block 86], pp. 62-63, p.238) showed that with the
assumption of transitivity this vast number  of learning steps reduces
to a mere $377$ steps.  We present a mathematical analysis of the
complexity of  Block's algorithm.  The algorithm has a complexity of order
$n^2$ given $n$ rules.  In addition, we improve 
Block's results exponentially, by introducing an algorithm that has
complexity of order less than $n \log n$.    

\end{abstract}


\pagebreak
\tableofcontents

\section{Introduction}

Language learning is thought to be a highly complex process. One of the
hurdles in learning a language is to learn the rules of syntax of the
language. Rules of syntax are often ordered in that before one can apply
one rule one must apply another. It has been thought that to learn the
order of $n$ rules one must go through all $n!$ permutations.  Thus to
learn the order of $27$ rules would require $27!$ steps or
$1.08889x10^{28}$ steps.  This number is much greater than the number of seconds since
the beginning of the universe!  In a brilliant analysis the  linguist
Block ([Block 86], pp. 62-63, p.238) showed that with the assumption of
transitivity this vast number  of learning steps reduces to a mere $377$
steps. 

We present a mathematical analysis of the complexity of  Block's
algorithm.  The algorithm has a complexity of order $n^2$ given $n$ rules.
In addition, we improve on Block's algorithm, by introducing an
algorithm that has complexity of order less than $n \log n$.  For
example, given $27$ rules our method requires $104$ steps.

\section{Block's algorithm} 

Given $n$ rules $R_1 \ldots R_n$ we are to
guess (learn) an unknown ordering of these rules.   Starting with any
given rule $Y$, we are to position a rule $X$. To do this we test
(query, learn) if $X$ is before $Y$,  if so we place it before $Y$  to
give the sequence $X Y$, else we place it after $Y$ resulting in $YX$.  
Now, assume we have ordered $i$ rules $Y_1 \ldots Y_i$ and we have to a
new place rule $X$, Block suggests we test for each $j$ if  $X < Y_j$,
if so all we need do is to place $X$ before $Y_j$ in our sequence $Y_1
\ldots Y_j \ldots Y_i$ to give $Y_1 \ldots X Y_j \ldots Y_i$.  The reason we  can do this
is because the rules are linearly ordered and hence by transitivity we
know $X$ is less than all rules beyond $Y_j$. Hence, we need  test no
further.  If $X$ is not less than $Y_j$ we continue until we find such a
$j$.  If we don't find such a $j$, then we know that $X$ is greater than
all all the given rules and we place $X$ at the end to give $Y_1 \ldots
Y_i X$.  Thus starting with one rule we gradually build up our 
knowledge about the order of the rules. \vspace{1pt}

\subsection{Complexity of Block's algorithm}

At any stage, given we know the order of rules $Y_1 \ldots Y_i$ , it
takes at most $i$ comparisons to place our new rule $X$.   Hence, to
learn the total rule set $R_1 \ldots R_n$ it takes at most $1+ 2 +
\ldots + i + \ldots + n$ steps.  This  worst case  is how computer
scientists measure the complexity of an algorithm.  

\begin{equation}
	 S(n) = \sum_{i=1}^n i -1  = \sum_{i=2}^n i  
\end{equation}

We subtract $1$ (or equivalently start with $2$) because the first rule
does not have to be tested or compared.   Thus, we have a measure of the
 complexity of Block's algorithm in defined in terms of the number of
 queries or steps needed to order $n$ rules.

\subsection{A simpler formula measuring Block's algorithm's complexity}

There is a simpler and shorter formula that eliminates the need to sum
up the steps and thereby can easily be done on a calculator.  It gives
an exact characterization of the steps needed to order $n$ rules:


\textbf{Exact complexity of Block's algorithm}
\begin{equation}
	 S(n) = \frac{(n^2 -n)}{2} + n -1 
\end{equation}

Since, the exponent of $2$ ultimately overpowers any addition or division by a constant, the complexity of Block's algorithm 
is $O(n^2)$.

For example, for $27$ rules it takes at most 
$ \sum_{i=1}^{27} i = 27 +	26 + \ldots + 2 = 377$  This is, of course, a very modest number when 
compared to the estimates of $27! = 1.08889x10^{28}$ steps by some
linguists.  For practical, purposes one would think this is enough and
we should be satisfied with such an astounding reduction of steps needed
to learn some $27$ rules.  

However, if we increase the number of rules
to hundreds, we quickly face numbers that may again seem insurmountable
for the developing brain of a child.  The method we present here
improves on Blocks algorithm significantly since it is exponentially
more efficient.  As a result, even the ordering of thousands of rules can be
learned relatively quickly.

For example, for one thousand rules, the number of steps required to
determine the ordering of the rules using Block's algorithm is 
$500,499$ steps. Below we will show that for this case we can improve
this to less than $8,977$ steps which is about 55 times faster.

\section{A fast algorithm to learn the order of $n$ rules}
Block's algorithm moves linearly from start to end to find and test if the rule is before or after the given rule.
Instead we will adapt binary search to start at the middle of the sequence of rules	and test the rule there.  This will 
give us an exponential improvement over Block's method.  

\subsection{A faster rule ordering algorithm}

Given $n$ rules and a rule $X$ that needs to be placed in its proper
order,  choose the middle rule, call it $Y$. This can be done since the
rules are linearly ordered.   Test to see if the rule is applied before
$Y$ if so test the middle rule between $Y$ and the first rule, else test
the middle rule between the end rule and rule $Y$.  Repeat until done.

\subsection{Complexity of the fast rule ordering algorithm}

In effect we use binary search to find the proper place to insert our
new rule $Y$ given $n$ rules. And, this process takes at most $n \log n$ steps. 
Hence, given we have $n$ rules to insert into at most $n$
rules, we have at most $n \log n$ steps to find the completely
ordered sequence of $n$ rules. 

Actually, the process takes fewer steps.   Since, work incrementally
starting with $n = 1$, call this one rule $Y$ and one rule $X$ to insert
we just have to make one test to see if $X < Y$ or not.  If it is less
than $Y$ we insert $X$ before $Y$ giving $X Y$ else we insert it after
$Y$ resulting in the partial sequence $Y X$.  We then pick the next rule
$Z$ and insert it. Given a sequence of $n$ rules we insert $X$ at each
stage according to the binary search algorithm above.  

Hence, for $n$ rules we require at most 
\begin{equation}
	B(n) = \lceil\log(n)\rceil + \lceil\log(n - 1)\rceil + \lceil\log(n - 2)\rceil + \ldots + \lceil\log(n - (n - 1))\rceil
\end{equation}

steps to order them. 

\textbf{Notation} Throughout this article we use $\log$ to mean the
logarithm base $2$. The notation, $\lceil\alpha\rceil$ denotes the
smallest integer $i$ greater than or equal to the number $\alpha$.  The
log of a number will in general not be an integer.  This promotes a real
(floating point)  number to the next integer since we are counting 
steps in an algorithm and they are discrete.

Expressed more succinctly we have:
\vspace{1pt}

\textbf{Complexity of the fast algorithm}

\begin{equation}	
	B(n) = \sum_{i=0}^{n-1} \lceil\log(n - i)\rceil  
		\label{BinaryX}	
\end{equation}

Note, that $B(n) < n \log n$.  Thus, the complexity is less than $O(n log n)$.

\subsection{An shorter but less accurate complexity function}

Looking at it from a different perspective, we can shorten the
description of the complexity function by using factorial.  

\begin{eqnarray*}
  	 \sum_{i=0}^{n-1}\log(n - i) = \log(n) + \log(n - 1) + \log(n - 2) + \ldots +  \log(n - (n - 1)) \\ 
   	=  \log( n(n-1)(n-2) \ldots (n-(n-1)) )  =  \log(n!)
\end{eqnarray*}

Thus, we can get an approximation to the function $B(n)$ in equation \ref{BinaryX} using the following:

\begin{equation}
	B_{f}(n) = \lceil\log( n! )\rceil
\end{equation} 

However, this formulation 
has the disadvantage that the size of the factorial quickly expands to
be computationally impractical.  In addition, the result is not as
accurate in that it underestimates the complexity because the $log$ is
used instead of $\lceil\log\rceil$ when $\log(n!)$ is calculated.  If we
add $n$ to the result, giving 	$\log(n!) + n$ we get an overestimate.
Thus, 

\begin{equation}
log(n!) < B(n) < log(n!) + n
\end{equation}

\section{Examples comparing the performance of the algorithms}

For example, for $27$ rules Block's algorithm requires $377$ steps. We
need only $104$ steps which is which is 3.6 times faster.  However, since there
is an exponential improvement using the fast algorithm,  the
differences  become more dramatic the larger the number of rules that
are to be ordered.  For example, for $1,000$ rules, some linguists would
tell us we need to search through $1000!$ possible combinations.  Using
Block's algorithm, we need only  $500,499$ steps, a much much smaller number. With
our improved fast algorithm we need less than $8,977$ steps which is more than 55
times faster (smaller) in this case. 
So, for instance, if a child would learn to order two rules per day (into its given set of learned rules), 
it would take 685 years to learn the ordering of 1,000 rules
using Block's method.   With our fast method it would take about 12 years.  

There is the question to what extent such considerations are relevant to learning a language.  
The above presupposed that the rules are linearly ordered.  
That means each rule stands in a unique linear relationship to all the other rules.  Partially ordered rules are yet another matter.

\pagebreak

\end{document}